\documentclass{article}
\usepackage{spconf,amsmath,graphicx,amsfonts,amssymb}
\usepackage{color,soul}
\usepackage{multirow,booktabs,diagbox}
\usepackage{url}
\usepackage{setspace}
\title{ChangeChat:  An Interactive Model for Remote Sensing Change Analysis via Multimodal Instruction Tuning}
\name{Pei~Deng, Wenqian~Zhou, Hanlin~Wu}
\address{School of Information Science and Technology, Beijing Foreign Studies University, Beijing, China}
\begin{document}
%
\maketitle
\begin{abstract}
    Remote sensing (RS) change analysis is vital for monitoring Earth's dynamic processes by detecting alterations in images over time. Traditional change detection excels at identifying pixel-level changes but lacks the ability to contextualize these alterations. While recent advancements in change captioning offer natural language descriptions of changes, they do not support interactive, user-specific queries. To address these limitations, we introduce ChangeChat, the first bitemporal vision-language model (VLM) designed specifically for RS change analysis. ChangeChat utilizes multimodal instruction tuning, allowing it to handle complex queries such as change captioning, category-specific quantification, and change localization. To enhance the model’s performance, we developed the ChangeChat-87k dataset, which was generated using a combination of rule-based methods and GPT-assisted techniques. Experiments show that ChangeChat offers a comprehensive, interactive solution for RS change analysis, achieving performance comparable to or even better than state-of-the-art (SOTA) methods on specific tasks, and significantly surpassing the latest general-domain model, GPT-4. Code and pre-trained weights are available at \url{https://github.com/hanlinwu/ChangeChat}.
\end{abstract}
\section{Introduction}
\label{sec:intro}

\begin{figure*}[t]
  \centering
  \includegraphics[width=\textwidth]{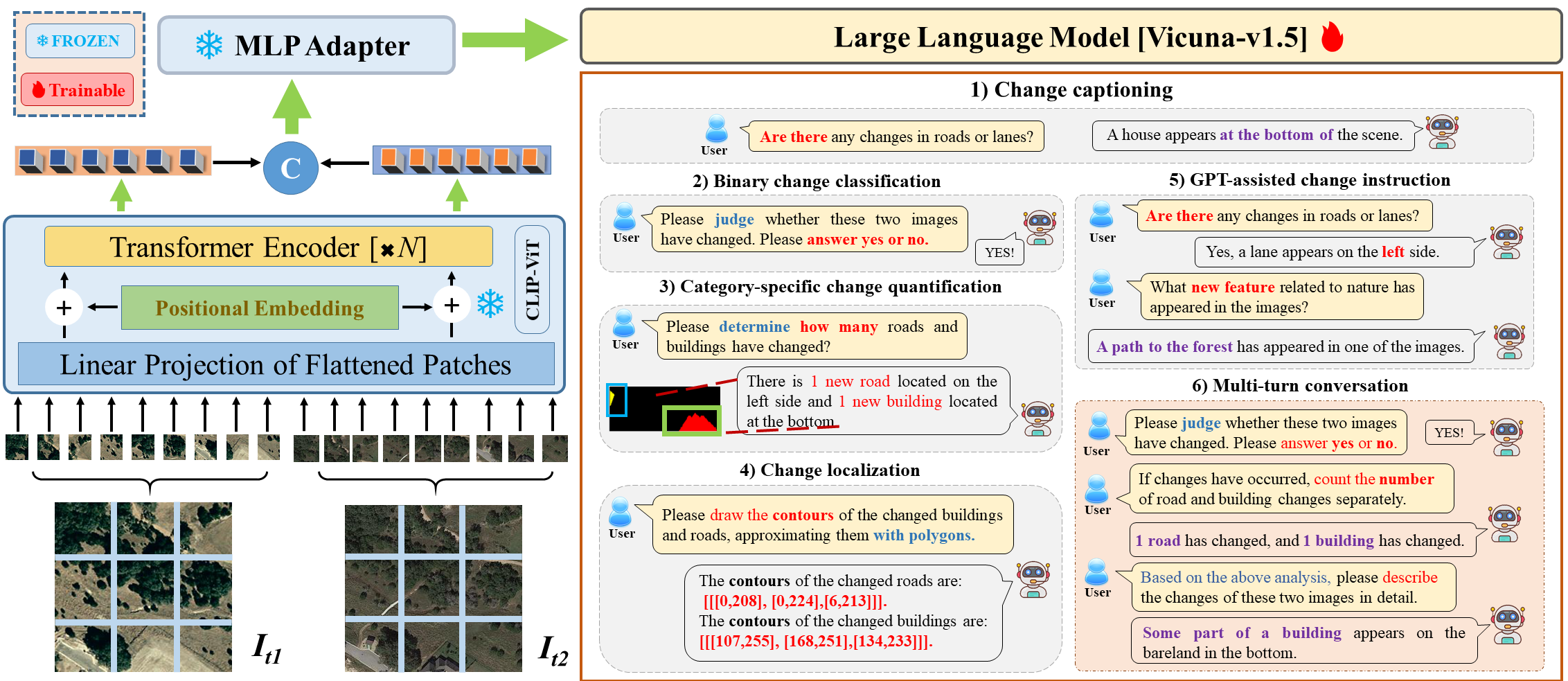}
  \caption{Overview of the proposed ChangeChat. The left side illustrates the network architecture, while the right side shows examples of various types of change analysis.}
\end{figure*}

Remote sensing (RS) change analysis is crucial for monitoring dynamic Earth processes, focusing on detecting alterations in images captured at different times over the same geographical area. This technique is fundamental to applications such as environmental monitoring \cite{jiao2024application} and disaster management \cite{al2024integrating}.

Change detection, traditionally the cornerstone of RS analysis, has excelled in identifying surface alterations at the pixel level \cite{bao2020ppcnet, zheng2022changemask}. Despite its accuracy, change detection often lacks the ability to contextualize the changes, leaving out details about the characteristics of the objects involved or their spatial relationships. To bridge this gap, ``change captioning'' has emerged, translating detected changes into natural language descriptions to provide a richer understanding of the observed changes \cite{liu2022remote,liu2023decoupling}. However, while enhancing interpretation, change captioning does not support interactive queries or offer user-specific information.

To address these limitations, the interactive Change-Agent model \cite{liu2024change} introduced multi-task learning to simultaneously handle both change detection and captioning. However, its reliance on an external large language model (LLM) and lack of an end-to-end framework restrict its functionality within predefined parameters. Recent vision-language models (VLMs) such as GPT-4 \cite{achiam2023gpt} and LLaVA \cite{liu2024visual} show promise across general domains but struggle with the unique characteristics of RS data, often leading to inaccurate or misleading interpretations.

To overcome these challenges, we introduce ChangeChat, the first bitemporal VLM designed for RS change analysis through multimodal instruction tuning. ChangeChat goes beyond predefined workflows, offering a flexible, interactive platform capable of responding to a diverse range of change-related queries. This capability extends to detailed tasks like change captioning, category-specific quantification, and precise change localization.

Despite these advances, the absence of specialized multimodal instruction-tuning datasets for RS change analysis initially limited ChangeChat's potential. In response, we developed the ChangeChat-87k dataset, comprising 87,195 instructions tailored specifically for RS change scenarios. For dataset generation, we first utilized the captions and change maps from the LEVIR-MCI dataset to obtain diverse multimodal change-related instructions through a rule-based automated pipeline. Additionally, inspired by recent advancements in instruction tuning \cite{liu2024visual}, we leveraged ChatGPT's in-context learning capabilities to automatically convert multimodal samples into appropriate instruction-following formats. This approach ensures a broad variety of instructions, enhancing both the flexibility and robustness of the dataset for RS change analysis tasks.


In summary, this work makes the following contributions:

\textbf{1) RS change instruction dataset:} We developed the RS change instruction dataset with 87k instruction-response pairs, using a pipeline that combines rule-based methods with GPT-assisted generation to automatically create instruction-response pairs. This dataset encompasses a variety of instruction types, including classification, counting, and descriptive tasks. It effectively addresses the shortage of instruction-following data in the field of RS change analysis and significantly enhances the performance of the ChangeChat model.

\textbf{2) ChangeChat:} We introduce ChangeChat, the first bitemporal VLM specifically designed for interactive RS change analysis. The proposed model integrates multimodal instruction tuning, creating a general-purpose RS change analysis assistant. It is capable of responding interactively to a broad array of queries, thereby providing a comprehensive solution that extends well beyond the capabilities of traditional change detection and captioning methods.

 
\section{Methodology}
\label{sec:methodology}

In this section, we first provide a detailed explanation of the construction of the instruction-tuning dataset for change analysis. Then, we introduce the architecture of ChangeChat.

\subsection{RS Change Instruction Dataset Generation}

The LEVIR-CC \cite{liu2022remote} has been a pioneering large-scale dataset for addressing the challenge of change captioning in remote sensing images (RSIs), comprising 10,077 pairs of bitemporal images, each accompanied by five natural language descriptions. However, due to the lack of fine-grained change information in LEVIR-CC, such as the quantity and location of changed objects, it is limited to the single task of change captioning. The subsequently released LEVIR-MCI \cite{liu2024change} dataset includes pixel-level change maps, but it is still insufficient for building a comprehensive model capable of in-depth change analysis. To address these limitations, we develop a new large-scale multimodal RS change instruction dataset tailored for change analysis tasks.

Inspired by the success of GPT-based automatic dataset generation tasks \cite{liu2024visual}, our RS change instruction dataset is constructed using a hybrid approach that combines rule-based methods with ChatGPT's in-context learning capabilities to automatically generate both instructions and responses. The proposed change instruction dataset contains six instruction types: 1) change description, 2) binary change detection, 3) category-specific change quantification, 4) change localization, 5) GPT-assisted change instruction, and 6) multi-turn conversation. Table \ref{tab:overview-dataset} summarizes the number of instructions for each type.

\textbf{Change captioning.} We first expand the change captioning sample ($I_{t1}$, $I_{t2}$, $C$) into an instruction-following version in a straightforward manner, \texttt{Human:\,$I_{t1}\,I_{t2}\,Q$\,<STOP> Assistant:\,$C$\,<STOP>}. Here, $I_{t1}$, $I_{t2}$ and $C$ represent the bitemporal images and the corresponding change caption, respectively. In this case, we fix the instruction $Q$ as: ``\textit{Please briefly describe the changes in these two images.}''

\textbf{Binary change classification.} We create instructions that ask ChangeChat to determine whether changes have occurred between the two images, expecting a binary ``\textit{yes}'' or ``\textit{no}'' answer. The ground truth is obtained based on the change map from the LEVIR-MCI dataset.

\textbf{Category-specific change quantification.} We create instructions that guide ChangeChat to quantify changes within specific categories like counting added buildings or roads. These quantity-related instructions are generated based on templates, and the ground truth for the responses is obtained by analyzing the change map using the OpenCV library.

\textbf{Change localization.} For precise change localization tasks, we create instructions that ask ChangeChat to delineate the contours of changed buildings and roads. The ground truth for the responses is obtained by extracting contours and approximating them with polygons from the objects in the change map. To adapt to the language model, we represent the polygons using a sequence of vertex coordinates: $P = \left[(x_1, y_1), (x_2, y_2), \cdots, (x_n, y_n)\right]$, 
where $(x_k,y_k)$ represents the normalized coordinates of the $k$-th vertex, and $n$ is the total number of vertices.

\textbf{GPT-assisted change instruction.} We leveraged ChatGPT's in-context learning capabilities to generate a wider variety of instruction-following data. We began by providing it with a system message to guide its responses. Then, we manually designed a few seed examples for each type of task to help it understand the desired output structure. Specifically, it generated two types of conversational data: (1) question-answer pairs based on five given captions describing changes, and (2) more fine-grained queries incorporating extracted contour and quantification information, enabling detailed questions about quantity and relative localization.

\textbf{Multi-turn conversation.} We designed multi-turn dialogues to encourage ChangeChat to perform change analysis using a chain-of-thought (CoT) approach. The instructions are presented in increasing difficulty, beginning with simple binary change classification, followed by change object and quantity identification, and progressing to the complex and detailed change captioning task.



\begin{table}[t]
\centering\small
\setlength{\tabcolsep}{3pt} 
\renewcommand{\arraystretch}{0.85}
\caption{Overview of the RS change instruction dataset.}
\label{tab:overview-dataset}
\begin{tabular}{lcc}
\toprule[1.5pt]
Instruction                            & Type           & Number                  \\ \midrule
Change captioning                     & Rule-based     & 34,075                  \\ 
Binary change classification                & Rule-based     & 6,815                   \\ 
Category-specific change quantification      & Rule-based     & 6,815                   \\ 
Change localization                      & Rule-based     & 6,815                   \\ 
GPT-assisted instruction             & GPT-assisted   & 26,600                  \\ 
Multi-turn conversation                & Rule-based     & 6,815                   \\ \midrule
Total                                  &                & 87,195                  \\ 
\bottomrule[1.5pt]
\end{tabular}
\end{table}

\subsection{Instruction Tuning for ChangeChat}
\subsubsection{ChangeChat Architecture}

ChangeChat follows an architecture inspired by LLaVA \cite{liu2024improved} but with significant adaptations to suit our change analysis tasks. The model is built upon three key components: i) a vision tower based on CLIP-ViT \cite{tay2017learning}, ii) a cross-modal adaptor, and iii) an LLM based on Vicuna-v1.5 \cite{zheng2024judging}. 

Our ChangeChat differs significantly from LLava in three aspects. First, LLava can only accept a single image as visual input and lacks the capability to conduct change analysis on multi-temporal images. Second, we enable ChangeChat to output spatial locations represented by coordinates through a carefully constructed instruction dataset, thus bridging the gap in language models' ability to handle visual localization tasks. Finally, LLava is oriented towards general domains and performs poorly in reasoning about RSIs. Next, we will provide a detailed introduction to each component.

\textbf{Vision tower.} We used the pre-trained CLIP-ViT \cite{tay2017learning} model as the visual feature extractor, which divides each $256 \times 256$ image into a $14 \times 14$ grid of patches and encodes each patch into a 1024-dimensional token.

\textbf{Cross-modal adaptor.} Following \cite{liu2024improved}, we employ a multi-layer perceptron (MLP) with one hidden layer to align the visual and language modalities. Specifically, the $1024$-dimensional tokens from the vision tower are mapped to $4096$ dimensions to serve as the input embeddings for the LLM.

\textbf{LLM.} We use the LLM Vicuna-v1.5 \cite{zheng2024judging} as the brain of our ChangeChat. We improved the Vicuna model by integrating visual embeddings from two different temporal phases, enhancing its capabilities in change analysis tasks. By fine-tuning the LLM with low-rank adaptation (LoRA) \cite{hu2021lora}, we optimize key matrix elements to enhance speed while preserving the model's linguistic capabilities. This approach integrates a broader range of contextual knowledge into change analysis tasks, enhancing our model's capabilities in change detection, counting, and localization.


\subsubsection{Training Details}
We initialize the model's weights using the pre-trained CLIP-ViT \cite{tay2017learning}, a pre-trained MLP from \cite{liu2024improved}, and Vicuna-v1.5 \cite{zheng2024judging} as the LLM. The LLM is efficiently fine-tuned using LoRA \cite{hu2021lora}, while the parameters of the visual tower and the MLP adapter are frozen. The LoRA fine-tuning is implemented with a rank $r$ of $64$ and a scaling factor $\alpha$ of $128$. The training leverages mixed-precision techniques to accelerate the process and reduce memory usage, while dynamic loss scaling is employed to maintain numerical stability.

We train ChangeChat for one epoch on the ChangeChat-87k dataset using the AdamW optimizer with a cosine learning rate scheduler. The learning rate is set to $2\times 10^{-4}$, with a warmup ratio of $3\%$. The training was completed using a single NVIDIA L20 GPU with 48GB of memory. To achieve an effective global batch size of $96$, we use a batch size of $16$ and accumulate gradients over $6$ steps.

\section{Experimental results}
\label{sec:experiments}

\subsection{Dataset and Evaluation Metrics}
We evaluate the model performance using 1929 samples from the LEVIR-CC test set. For the change captioning task, we employ BLEU-1\cite{papineni2002bleu} for precision between generated hypotheses and reference sentences, METEOR \cite{banerjee2005meteor} for alignment consideration of synonyms and word order, ROUGE-L \cite{lin2004rouge} for recall based on the longest common subsequence, and CIDEr \cite{vedantam2015cider} for similarity with multiple references. Additionally, we assessed accuracy and recall for binary change classification, and used mean absolute error (MAE) for category-specific change quantification.

\subsection{Comparison with SOTA Change Captioning Models}
In this section, we evaluate the performance of ChangeChat across various change analysis tasks. For the change captioning task, we compare it with SOTA methods, including Capt-Dual-Att \cite{park2019robust}, DUDA \cite{park2019robust}, MCCFormer-S \cite{qiu2021describing}, MCCFormer-D \cite{qiu2021describing}, RSICCFormer \cite{liu2022remote}, Prompt-CC \cite{liu2023decoupling}.

To achieve more robust change captioning results, we propose a CoT reasoning approach, guiding ChangeChat through step-by-step reasoning in a multi-turn dialogue. Specifically, we first prompt ChangeChat to determine if any changes have occurred, then guide it to assess specific changes in roads and buildings. Finally, it generates the change caption based on the gathered information.

The results are shown in Table \ref{tab:rslt-cc}, ChangChat achieves comparable or even better results than SOTA models. Notably, all the comparison models are limited to the single task of change captioning, while our proposal is a more general, multi-task-oriented change analysis system.

\begin{table}[ht]
  \centering\small
  \setlength{\tabcolsep}{3pt}
  \renewcommand{\arraystretch}{0.85}
  \vspace{-0.5em}
  \caption{Comparison with SOTA methods on the change captioning task. The best is in \textbf{bold}, the second best is \underline{underlined}.}
  \begin{tabular}{lcccc}
    \toprule[1.5pt]
    Method           & BLEU-1 & METEOR & ROUGE-L        & CIDEr-D         \\
    \midrule
    Capt-Dual-Att \cite{park2019robust}   & 78.17  & 35.23  & 71.60           & 127.51          \\
    DUDA \cite{park2019robust}            & 79.64  & 35.76  & 71.47          & 128.24          \\
    MCCFormer-S \cite{qiu2021describing}     & 79.38  & 36.88  & 71.06          & 127.90           \\
    MCCFormer-D \cite{qiu2021describing}     & 76.23  & 35.18  & 68.43          & 121.10           \\
    RSICCFormer \cite{liu2022remote}     & 81.96  & 38.16  & 72.57          & 132.00             \\
    PSNet \cite{liu2023progressive}      &  81.97  & 37.92   &  73.10   &   132.87  \\
    Prompt-CC \cite{liu2023decoupling}       & \textbf{83.66}  & \textbf{38.82}  & \underline{73.72}          & \underline{136.44}          \\
    ChangeChat(Ours) & \underline{83.14}  & \underline{38.73}  & \textbf{74.01} & \textbf{136.56} \\
    \bottomrule[1.5pt]
  \end{tabular}%
  \label{tab:rslt-cc}%
  \vspace{-1em}
\end{table}%

\begin{table}[t]
\centering\small
\setlength{\tabcolsep}{2pt}
\renewcommand{\arraystretch}{0.85}
\vspace{-0.5em}
\caption{Comparison on binary change classification and change quantification tasks.}
\label{tab:gpt4}
\resizebox{\columnwidth}{!}{
\begin{tabular}{c c c c c}
\toprule[1.5pt]
\multirow{2}{*}{Model} & \multicolumn{2}{c}{{Binary Classification}} & \multicolumn{2}{c}{{Change Quantification}} \\
\cmidrule(lr){2-3} \cmidrule(lr){4-5}
 & {Acc.} & {Recall} & {MAE (road)} & {MAE (building)} \\
\midrule
GPT-4 & 84.81\% & 86.62\% & 0.62 & 2.91 \\
ChangeChat & \textbf{93.21\%} & \textbf{92.53\%} & \textbf{0.33} & \textbf{2.67} \\
\bottomrule[1.5pt]
\end{tabular}}
\end{table}

\subsection{Evaluation on Diverse Change Analysis Tasks}
For more diverse change analysis tasks, due to the lack of prior research, we take the GPT-4 \cite{achiam2023gpt} as the baseline.

\textbf{Binary change classification.} We design an instruction for ChangeChat and GPT-4 to determine whether a change has occurred between two RSIs, prompting it to respond with only \textit{yes} or \textit{no} using the phrase "\textit{Please answer yes or no}". As shown in Table \ref{tab:gpt4}, ChangeChat outperforms GPT-4 in both accuracy and recall.

\textbf{Category-specific change quantification.} We focus on quantifying changes in roads and buildings, converting number-related outputs from the language model into Arabic numerals using predefined rules to calculate MAE with the ground truth. Table \ref{tab:gpt4} demonstrates that our method outperforms SOTA large VLMs in this task.

\begin{figure}[t]
\centering
\includegraphics[width=\linewidth]{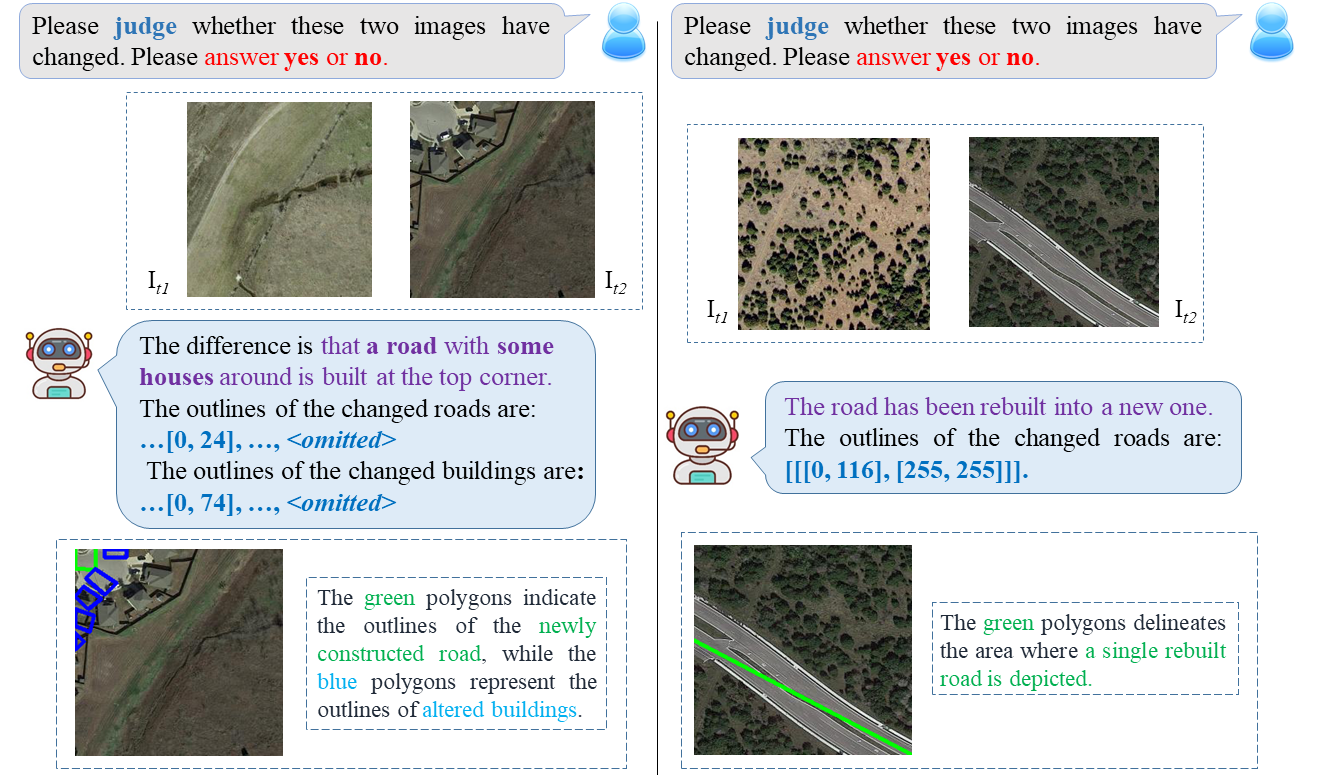}
\caption{Two examples of change localization are provided, with the generated coordinates visualized.}
\label{fig:change_location}
\end{figure}

\textbf{Change localization} To improve reasoning efficiency, we use highly simplified polygons to approximate the general location of the changed objects. We present two examples of change localization in Fig.\,\ref{fig:change_location}. The results show that ChangeChat can effectively identify and localize the changed objects.

\subsection{Discussion on CoT reasoning}
To validate the effectiveness of our CoT reasoning strategy, we compare the change captioning results with and without using CoT reasoning, as shown in Table \ref{tab:cot}. The results indicate that guiding ChangeChat through increasingly complex instructions indeed improves its performance.

\begin{table}[ht]
  \centering\small
  \renewcommand{\arraystretch}{0.85}
  \setlength{\tabcolsep}{4pt}
  \vspace{-0.5em}
  \caption{Comparison between with and w/o CoT reasoning.}
  \begin{tabular}{ccccc}
    \toprule[1.5pt]
    CoT Reasoning        & BLEU-1 & METEOR & ROUGE-L & CIDEr-D \\
    \midrule
    $\times$      & 81.16                              & 37.05                              & 73.37                              & 135.46                              \\
    $\checkmark$ & \textbf{83.14}                     & \textbf{38.73}                     & \textbf{74.01}                     & \textbf{136.56}                     \\
    \bottomrule[1.5pt]
  \end{tabular}%
  \vspace{-1em}
  \label{tab:cot}%
\end{table}%
\section{Conclusion}
\label{sec:conclusion}
In this paper, we introduced ChangeChat, the first bitemporal VLM designed specifically for RS change analysis. Unlike traditional change captioning models, ChangeChat supports interactive, user-specific queries through multimodal instruction tuning. The model, enhanced by the ChangeChat-87k dataset, excels in tasks such as change captioning, category-specific quantification, and change localization. Experiments show that ChangeChat outperforms SOTA methods and significantly exceeds the performance of the latest general-domain large model, GPT-4, demonstrating its effectiveness in comprehensive RS change analysis.

\twocolumn[\clearpage]

\bibliographystyle{IEEEbib}
\bibliography{refs}

\end{document}